\documentclass{article}

\usepackage{arxiv}

\usepackage[utf8]{inputenc} 
\usepackage[T1]{fontenc}    
\usepackage{hyperref}       
\usepackage{url}            
\usepackage{booktabs}       
\usepackage{amsfonts}       
\usepackage{nicefrac}       
\usepackage{microtype}      
\usepackage{amsmath}
\usepackage{cleveref}       
\usepackage{acronym}
\usepackage{graphicx}
\usepackage[numbers]{natbib}
\usepackage{doi}
\usepackage{rotating}
\usepackage{lscape}
\usepackage{array}     
\usepackage{caption}   

\acrodef{SOTA}[SOTA]{State-of-the-Art}
\acrodef{DL}[DL]{Deep Learning}
\acrodef{NLP}[NLP]{Natural Language Processing}
\acrodef{NN}[NN]{Neural Network}
\acrodef{CNN}[CNN]{Convolutional Neural Network}
\acrodef{NAS}[NAS]{Neural Architecture Search}
\acrodef{HW-NAS}[HW-NAS]{Hardware-aware Neural Architecture Search}
\acrodef{MCU}[MCU]{microcontroller}
\acrodef{ACO}[ACO]{Automatic Code Optimization}
\acrodef{AI}[AI]{Artificial Intelligence}
\acrodef{IC}[IC]{Image Classification}
\acrodef{SR}[SR]{Super Resolution}
\acrodef{NACOS}[NACOS]{Hardware Aware-Neural Architecture and Compiler Optimizations co-Search}

\title{Combining Neural Architecture Search and Automatic Code Optimization: A Survey}

\date{}

\author{
  \href{https://orcid.org/0009-0009-1313-073X}{\includegraphics[scale=0.06]{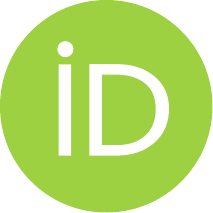}\hspace{1mm}Inas Bachiri}\thanks{Contact author} \\
  Ecole nationale Supérieure d'Informatique\\
  Algiers, Algeria \\
  \texttt{ji\_bachiri@esi.dz} \\
  \And
  \href{https://orcid.org/0000-0002-5259-0749}{\includegraphics[scale=0.06]{orcid.pdf}\hspace{1mm}Hadjer Benmeziane} \\
  IBM Research Europe\\
  Zurich, Switzerland\\
  \texttt{Hadjer.Benmeziane1@ibm.com} \\
  \And
  \href{https://orcid.org/0000-0002-9350-3998}{\includegraphics[scale=0.06]{orcid.pdf}\hspace{1mm}Riyadh Baghdadi} \\
  New York University Abu Dhabi\\
  Abu Dhabi, UAE\\
  \texttt{baghdadi@nyu.edu} \\
  \And
  \href{https://orcid.org/0000-0002-7550-484X}{\includegraphics[scale=0.06]{orcid.pdf}\hspace{1mm}Smail Niar} \\
  Université Polytechnique Hauts-de-France,\\
  LAMIH, INSA\\
  Valenciennes, France\\
  \texttt{Smail.Niar@uphf.fr} \\
  \And
  \href{https://orcid.org/0000-0002-7490-5350}{\includegraphics[scale=0.06]{orcid.pdf}\hspace{1mm}Hamza Ouarnoughi} \\
  Université Polytechnique Hauts-de-France,\\
  LAMIH, INSA\\
  Valenciennes, France\\
  \texttt{Hamza.Ouarnoughi@uphf.fr} \\
  \And
  \href{https://orcid.org/0000-0002-3042-1477}{\includegraphics[scale=0.06]{orcid.pdf}\hspace{1mm}Abdelkrime Aries} \\
  Ecole nationale Supérieure d'Informatique\\
  Algiers, Algeria\\
  \texttt{ab\_aries@esi.dz} \\
}

\renewcommand{\headeright}{Bachiri et al.}
\renewcommand{\undertitle}{}
\renewcommand{\shorttitle}{}

\hypersetup{
pdftitle={Combining Neural Architecture Search and Automatic Code Optimization: A Survey},
pdfsubject={hwnas, aco},
pdfauthor={Inas Bachiri},
pdfkeywords={Efficient Neural Networks, Hardware-aware Neural Architecture Search, Automatic Code Optimization, Compiler Auto Scheduler},
}

\begin{document}
\maketitle

\begin{abstract}
	\ac{DL} models have experienced exponential growth in complexity and resource demands in recent years. Accelerating these models for efficient execution on resource-constrained devices has become more crucial than ever. Two notable techniques employed to achieve this goal are \ac{HW-NAS} and \ac{ACO}. \ac{HW-NAS} automatically designs accurate yet hardware-friendly \acp{NN}, while \ac{ACO} involves searching for the best compiler optimizations to apply on \acp{NN} for efficient mapping and inference on the target hardware. 
     This survey explores recent works that combine these two techniques within a single framework. We present the fundamental principles of both domains and demonstrate their sub-optimality when performed independently. We then investigate their integration into a joint optimization process that we call Hardware Aware-\textbf{N}eural \textbf{A}rchitecture and \textbf{C}ompiler \textbf{O}ptimizations co-\textbf{S}earch (NACOS). 
\end{abstract}


\section{Introduction}~\label{sec:intro}
    In the quest for achieving state-of-the-art performance, \ac{DL} researchers have been designing highly complex Neural Networks (\acp{NN}) that have revolutionized our lives. 
    However, these progressive improvements come with significant increases in \acp{NN}' size, complexity, energy consumption, and inference latency, making their deployment on resource-constrained devices challenging.
    Given the plethora of available hardware platforms, ranging from powerful clusters to tiny IoT devices, it is essential to design efficient and hardware-friendly \acp{NN}. 
    This can be achieved by intervening at three levels: \textit{model design, software stack, and hardware stack}~\citep{efficient-dl-survey}: 

    \begin{enumerate}
        \item Methods for improving {\bf{model design}} include compression techniques like Quantization~\citep{quantiz}, learning techniques like Distillation~\citep{knw-distill} and optimization techniques, such as Hardware-aware Neural Architecture Search (\ac{HW-NAS}). The latter is a multi-objective optimization process automating the design of \acp{NN} through maximization of accuracy and hardware efficiency metrics. 
        
        \item At the {\bf{software stack}} level, we can either optimize the implementation provided by \ac{DL} frameworks such as PyTorch~\citep{torch}, or the code optimization strategies employed by the compiler on the low-level code of \acp{NN}~\citep{tiramisu-autosched,autotvm}.
        
        \item  The {\bf{hardware level}} mainly includes designing specialized hardware and tuning accelerator configurations for training and running \ac{NN} models inference~\citep{sparsehardware}.
    \end{enumerate}

\begin{figure}
        \centering
        \includegraphics[width=.4\textwidth]{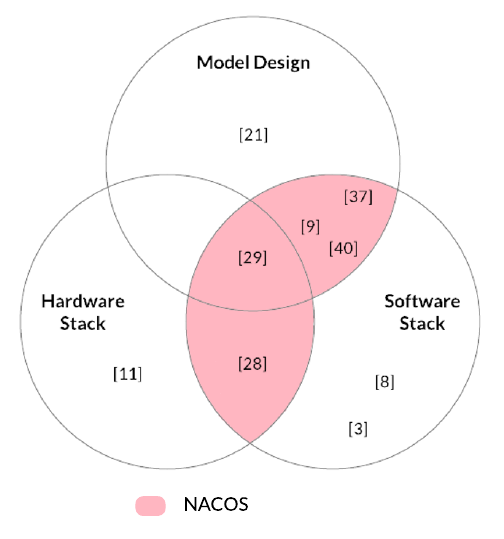}
        \caption{Cross-level joint deep learning optimization methods. In this paper, we explore works that combine HW-NAS and Automatic Code Optimization (NACOS) }
        \label{fig:cross-level}
    \end{figure}

    Each of these techniques contributes in optimizing \acp{NN}. 
    However, independent execution of these techniques leads to sub-optimal results.  For example, designing a network using \ac{HW-NAS}, and then applying quantization on top of it may lead to a drop in accuracy~\citep{quantiz}.  Therefore, combining techniques within and across the three design levels has been explored. For instance, at the model design level, NAS has been combined with Quantization~\citep{nas-with-quant}. Also, between the model design level and the hardware level, \ac{HW-NAS} has been combined with the exploration of hardware accelerator designs on FPGA~\citep{nas-with-hw}. Finally, between the model design level and the software stack level, compression has been combined with compiler optimizations~\citep{CoCoPIE}, illustrating the potential of synergies. Figure~\ref{fig:cross-level} summarizes some cross-level works that have been published. 

    A promising yet under-explored design methodology is to jointly explore \ac{HW-NAS} and Automatic Code Optimization (\ac{ACO}). 
    For the sake of brevity, we give it the name of NACOS (\textbf{N}eural \textbf{A}rchitecture and \textbf{C}ompiler \textbf{O}ptimizations co-\textbf{S}earch). 
    
    In their hardware efficiency evaluation, \ac{HW-NAS} techniques consider the implementation provided by the \ac{DL} framework as it is. This gives a rough approximation of the latency and energy consumption before applying code optimizations, resulting in the risk of overlooking optimal architectures during the search. On the other hand, \ac{ACO} usually works on a given network with a fixed architecture, regardless of its performance and optimality. Combining \ac{HW-NAS} and \ac{ACO} would allow an accurate approximation of the hardware efficiency metrics in \ac{HW-NAS} and an \ac{ACO} that is adapted to the best network. 

    This survey aims to explore NACOS methods and provide insights into the state-of-the-art methodologies by highlighting the following attributes:

    \begin{itemize}
    
        \item We first cover the fundamental principles of Hardware-aware Neural Architecture Search (\ac{HW-NAS}) and Automatic Code Optimization (\ac{ACO}) in Section~\ref{sec:background}.
    
        \item In Section~\ref{sec:motivation}, we highlight the drawbacks of performing each technique independently and motivate their joint optimization.
    
        \item We propose a taxonomy to classify NACOS methods based on their key components in Section~\ref{sec:taxonomy}. In addition, we provide a detailed overview of existing works with a focus on their search space design.
    
        \item We then discuss the exploration algorithms and the candidate evaluation strategies employed in existing methods in Section~\ref{sec:search}.
    
        \item In the last section, we address the challenges and limitations of the existing works and propose some research directions to advance this area further.
    
    \end{itemize}

\section{Background}~\label{sec:background}
    In this section, we provide a high-level overview of \ac{HW-NAS} and \ac{ACO} techniques.

    \subsection{Hardware-aware Neural Architecture Search}
        The end-to-end design process of \ac{NN} architectures proves challenging when performed manually, as it requires time and expertise. This is why many researchers resort to \ac{NAS}~\citep{NAS-survey}, which is a set of techniques for automatically designing \ac{NN} architectures. Formulated as an optimization problem, \ac{NAS} searches for the best architecture within a set of \ac{NN} candidates, referred to as the \textbf{search space}, using a \textbf{search algorithm} that optimizes one objective, such as accuracy, or multiple objectives, such as execution time, energy consumption, memory usage, etc. Due to the time-consuming training of every sampled architecture, an \textbf{evaluation strategy} is used to estimate the objective(s) with minimal or no training.

        \ac{NAS} has been successful in many tasks, including Image Classification~\citep{proxylessnas}, object detection~\citep{nas-obj}, and keyword spotting~\citep{autokws}. However, following the \ac{DL} trend, the found models require intensive compute power and memory resources due to their complex nature. This complexity also makes it hard to deploy them to hardware for efficient inference. This challenge has spurred the development of novel multi-objective \ac{NAS} methods, known as Hardware-aware Neural Architecture Search (HW-NAS) ~\citep{HadjerArxiv}. \ac{HW-NAS} is described with the same components as \ac{NAS}. The search space can be tailored to the target hardware and task~\citep{hurricane}, and the search algorithm finds the Pareto front, which represents the best architectures in terms of trade-off between the considered objectives. These objectives include hardware metrics such as latency~\citep{fbnet,proxylessnas}, FLOPs~\citep{mnasnet,efficientnet}, energy consumption~\citep{energy-nas} and so on. The complexity of \ac{HW-NAS} is hindered by the evaluation of these objectives, which requires estimation techniques such as using lookup tables~\citep{fbnet}, analytical approximations~\citep{nascaps}, and surrogate models~\citep{proxylessnas}. For in-depth details about \ac{HW-NAS} techniques, we refer readers to~\citep{hwnas-survey}.

    \subsection{Automatic Code Optimization}

        Automatic code optimization (ACO), also known as auto-scheduling, refers to a set of techniques that are implemented at the compiler level with the aim of automating the optimization of high-level code. ACO is achieved by identifying the most effective schedule, which is essentially a sequence of selected code transformations such as parallelization, loop tiling, interchange, and vectorization, applied under a specific order with specific parameters. This schedule enhances the efficiency of the code's implementation and hardware mapping without altering its semantics.

        One example is the Halide Auto-scheduler~\citep{halide}, which automatically explores the best schedule for Halide programs. The search space includes various code transformations like parallelization, loop unrolling, interchange, fusion, and tiling. This exploration is performed using Beam Search, guided by a \ac{NN}-based cost model predicting the speedup achievable with a particular schedule, given manually engineered features representing both the program and the schedule.

        On the other hand, to automate finding the optimal sequence of transformations to apply on a Tiramisu~\citep{tiramisu} program, a \ac{DL}-based auto-scheduler~\citep{tiramisu-autosched} has been implemented. The search process is conducted using both Beam Search and Monte Carlo Tree Search, leveraging a cost model as a speedup estimator to navigate the space. The cost model predicts the speedup that we’d get from applying a schedule on the program, given a set of automatically-extracted features.

        Another aspect of auto-scheduling involves searching for the best parameter configuration of a given schedule. Code transformations typically involve one or more parameters that need tuning. Choosing appropriate parameters is crucial to effectively apply the transformation and achieve desired results.  An example is AutoTVM~\citep{autotvm}— a compiler infrastructure tool implemented for the TVM~\citep{tvm} compiler that uses simulated annealing to search the parameter space, and leverages a \ac{NN} model to rank program versions with different transformation parameters based on high-level computation graph abstractions, which significantly optimizes tensor programs.

\section{Motivation}~\label{sec:motivation}

    Given a hardware platform with specific efficiency constraints like latency or memory usage specifications, a  classical \ac{HW-NAS} framework will search for the best possible \ac{NN} architecture that satisfies these constraints. This involves evaluating candidate \ac{NN} architectures under a fixed scheduling strategy from a deep learning library (e.g., TensorFlow or PyTorch). However, the discovered architecture may not be optimal for the target hardware when executed with a certain scheduling strategy.

\begin{figure}
        \begin{center}
            \includegraphics[width=0.5\columnwidth]{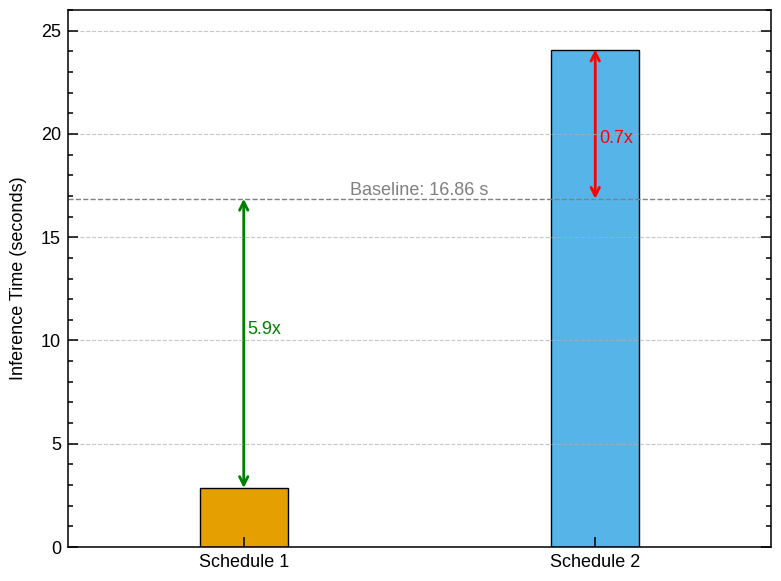}
        \end{center}
        \caption{The inference latency of VGG16 using two different compiler schedules, on 100 image samples from ImageNet~\citep{imagenet}. Schedule 1 and Schedule 2 are obtained by applying different sequences of parallelization, loop tiling, and fusion with various parameters using MLIR~\protect\citep{mlir}, and executing on an Intel Core i7 processor with 32 GB of RAM. The two schedules perform differently on the network; Schedule 1 outperforms Schedule 2 by making the inference on VGG16 faster. The values on the arrow lines represent the acceleration relative to the baseline time, with green indicating acceleration and red indicating deceleration. The baseline time represents the original inference time of VGG16 before applying the schedules.}
        \label{fig:exp1}
    \end{figure}

    As illustrated in Figure~\ref{fig:exp1}, we observe variations in inference latency values for the same \ac{NN} when different schedules are applied. For instance, applying two different schedules to VGG16~\citep{vgg} results in largely distinct latency values. This highlights the significance of the schedule choice in latency estimation, and the risk for the \ac{HW-NAS} process to overlook optimal architectures if it does not consider compiler optimizations. For example, if we were constrained by a maximum of 10s latency, and the fixed compiler schedule that we are using is schedule 2, the VGG16 architecture would be considered to not meet the constraint and will be overlooked, even though it could fit in the constraint really well if schedule 1 was used.

\begin{figure}
        \begin{center}
            \includegraphics[width=0.6\columnwidth]{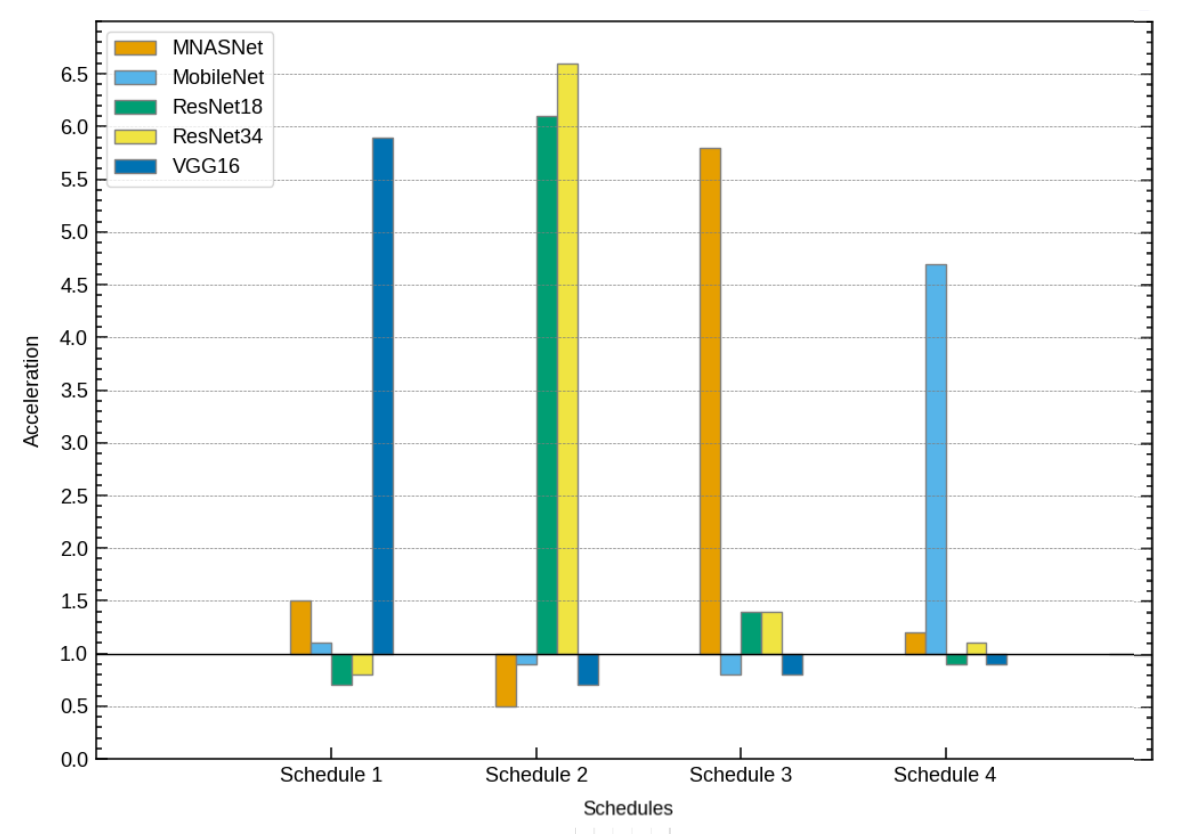} 
        \end{center}
        \caption{The accelerations of various neural networks under different scheduling strategies. Schedule 1 significantly speeds up VGG16, but does not improve and even worsens performance for other networks. Similarly, Schedule 2 optimizes ResNet18 and ResNet34 but slightly increases VGG16's latency.  This demonstrates that a well-optimized schedule for one neural network does not necessarily optimize other networks.}
        \label{fig:exp2}
    \end{figure}

    On the other hand, a scheduling strategy that optimizes performance for one \ac{NN} may not yield similar benefits for others. In Figure~\ref{fig:exp2}, Schedule 1 significantly reduces the inference latency of VGG16~\citep{vgg}, achieving a speed-up of approximately 5.9 times compared to the original execution latency. However, this same schedule does not universally optimize the performance across all networks. For networks like ResNet18, ResNet34, MobileNet~\citep{mobilenet}, and MNASNet~\citep{mnasnet}, it results in inference latencies that are comparable to or worse than their original latencies. Similarly, Schedule 2 shows a noticeable speed-up for ResNet18~\citep{resnet} and ResNet34. Yet, for other networks,  it may increase the execution latency slightly. This shows that searching for a good schedule, whether manually or through \ac{ACO}, can be sub-optimal if the \ac{NN} design is not considered in the process.

    Methods such as NACOS are thus primordial to explore both search spaces and come up with a design of \ac{NN} architectures and their associated schedules for optimal performance and hardware efficiency on a given task. 

\section{Taxonomy}~\label{sec:taxonomy}

\begin{figure*}
        \begin{center}
            \includegraphics[width=\textwidth]{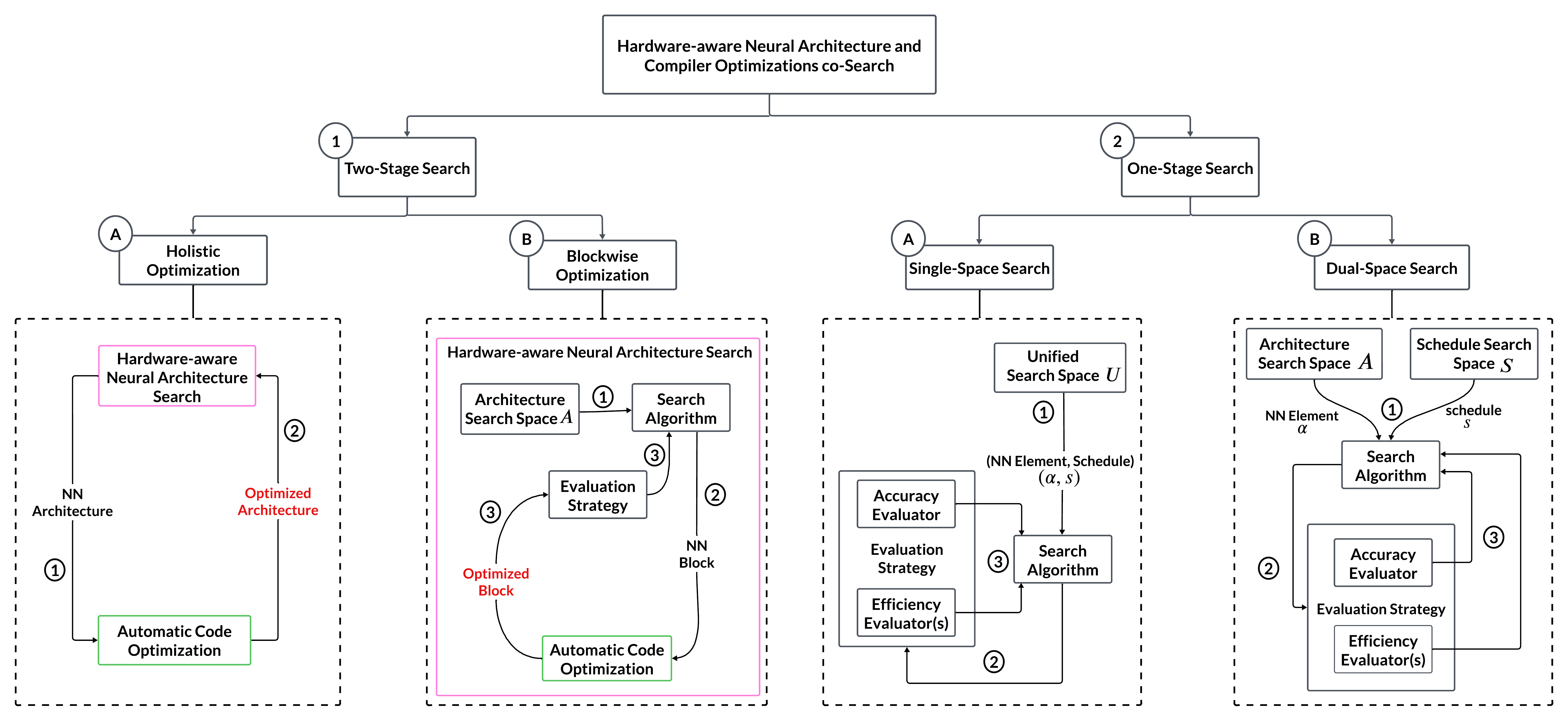}
        \end{center}
        \caption{Hardware Aware \textbf{N}eural \textbf{A}rchitecture and \textbf{C}ompiler \textbf{O}ptimizations co-\textbf{S}earch (NACOS) Taxonomy}
        \label{fig:taxonomy}
    \end{figure*}

    In this section, we propose a taxonomy to categorize NACOS techniques based on their search flow, providing a structured foundation for future work. Our taxonomy is summarized in  Figure~\ref{fig:taxonomy}. At a high level, we categorize NACOS methods into two classes: two-stage search methods and one-stage search methods. The first treats \ac{HW-NAS} and \ac{ACO} as separate search processes, while the second treats them as a single one. We describe these two classes of methods in the rest of this section. In addition, an overview of existing NACOS methods is presented in Table ~\ref{tab:methods}.

        \subsection{Two-stage Search}
        This class treats \ac{HW-NAS} and \ac{ACO} as separate stages, engaging in two distinct search processes that interactively contribute to the final outcome. For each sampled \ac{NN} architecture candidate in the \ac{HW-NAS} stage, the \ac{ACO} optimizes its code and provides the \ac{HW-NAS} with precise efficiency metrics. 
        Equation~\ref{eq:stage1} shows the optimization problem solved by the \ac{HW-NAS} stage. In this equation, $\alpha$ represents a candidate architecture from the architecture search space denoted as $A$. The search seeks to minimize the loss function of the neural network denoted as $\mathcal{L}$ and
        the cost of executing the neural network on hardware denoted as $\mathcal{C}$. Several metrics, such as latency, energy consumption, or memory footprint, can be combined in the evaluation of $\mathcal{C}$.

        \begin{equation}
            \min_{\alpha \in A} \mathcal{L}(\alpha), \mathcal{C}(\alpha)
            \label{eq:stage1}
        \end{equation}

        In the \ac{ACO} stage, the search for the best transformation sequence applied to $\alpha$ is formulated in Equation~\ref{eq:stage2}. Here, $s$ represents a schedule (a sequence of code optimizations) in the schedules space, denoted as $S$. The problem is then cast as a constrained optimization one. We seek to minimize the cost of executing the candidate architecture $\alpha$ on the hardware when $s$ is applied, $\mathcal{C}(s(\alpha))$, under the constraint of keeping the same loss or reducing it.
        
         \begin{equation}
            \begin{split}
        		& \min_{s\;\in\;S} \mathcal{C}(s(\alpha)) \\
        		&\text{subject to}\; \mathcal{L}(s(\alpha)) \le \mathcal{L}(\alpha)
        	\end{split}
        	\label{eq:stage2}
        \end{equation}

        Depending on the nature of the candidates given by the \ac{HW-NAS} stage, i.e. whether they are blocks or entire architectures, techniques within this category can be classified into:

            \subsubsection{Holistic Optimization}
                The \ac{NAS} stage within this category first explores the associated architecture space and discovers an entire \ac{NN} architecture $\alpha$. The resulting architecture $\alpha$ is then handed over to the \ac{ACO} stage, which optimizes its code,
                allowing the expansion of the search space to accommodate larger models with greater accuracy. Optimizing the \ac{NN} and handing it back to \ac{HW-NAS} allows it to find a more accurate model while still satisfying the efficiency constraints, making full use of the target hardware's resources. The process is illustrated in Figure~\ref{fig:taxonomy}(1.A).

                In this category, we find MCUNet~\citep{mcunet}, a model-software co-design framework for \acp{MCU}. This framework jointly designs an efficient \ac{NN} and its corresponding inference schedule to fit the tight memory resources on \acp{MCU}. This is achieved through two interactive components : TinyNAS and TinyEngine. TinyNAS is a \ac{HW-NAS} module that first prunes  the architecture search space, then performs one-shot search (refer to Section~\ref{sec:search}) in the optimized space for the best architecture that satisfies memory and latency constraints using the evolutionary algorithm.  The selected architecture by TinyNAS is handed to TinyEngine, a memory-efficient inference library that eliminates the unnecessary memory overhead and performs \ac{ACO}. TinyEngine optimizes the whole architecture and adapts its schedule to make full use of the limited resources in the \acp{MCU}, which will give TinyNAS more space to search for a larger, better performing architecture.
                        
        \subsubsection{Blockwise Optimization}
             In this category of appraoches, the \ac{ACO} stage searches for the best compiler schedule for each building block of the \ac{NN} architecture independently during the search process of \ac{NAS}, instead of the entire architecture altogether. As shown in Figure~\ref{fig:taxonomy} (1.B), at each iteration of the joint search process, the \ac{NN} block proposed by \ac{HW-NAS} is handed to the \ac{ACO} module for optimization. The optimized neural network blocks are subsequently used by \ac{HW-NAS} to construct the final architecture.

            An example of this category of techniques is CHaNAS~\citep{chanas}, a framework that co-searches for an \ac{NN} architecture and a corresponding compiler scheduling policy that maps the model onto the target hardware. To do this, it constructs a super-network~\citep{enas} to represent the architecture search space by stacking medium sized neural blocks, constituted of elastic MBConv cells. Such cells consist of a $1 \times 1$ convolution, a $k\times k$ depthwise convolution, a Squeeze and Excitation~\citep{se} block and another $1 \times 1$ convolution. All candidate \ac{NN} architectures can be extracted from this super-network, allowing for quick accuracy evaluation through weight sharing~\citep{weight-sharing-survey} (refer to Section~\ref{sec:search}). The framework follows a hierarchical exploration process centered around neural blocks, involving two coordinated search phases. First, the compiler scheduler goes through the super-network, transforms each block into a computational sub-graph and optimizes it for the target hardware. The optimization is done at two levels: graph-level optimization, where the block is optimized using layer fusion and data layout transformations, and then using lower-level optimizations where the best schedule for each operator is independently searched using an evolutionary-like heuristic with a Gaussian Process-based cost model. Then, the \ac{HW-NAS} component explores different sequences of the neural blocks (sub-networks of the super-network) that have already been optimized by \ac{ACO}. For this, the search space is automatically pruned according to the performance constraints. After that, an evolutionary algorithm is used to search for the solution in the reduced space, guided by a \ac{NN} accuracy predictor and a latency lookup table.

    \subsection{One-stage Search}
    In this category, neural architectures and compiler schedules are explored using one single search algorithm that simultaneously searches for both of them in each iteration. Equation~\ref{eq:one_stage} illustrates how the optimization process is cast. The exploration can take place in a unified search space, denoted as $U$, or in two distinct spaces $A$ and $S$

    \begin{equation}
         \min_{\substack{(\alpha, s) \in U \\ or \\ (\alpha, s) \in (A,S)}} \mathcal{L}(s(\alpha)), \mathcal{C}(s(\alpha))
        \label{eq:one_stage}
    \end{equation}
    
    Based on that, we can further classify techniques within this category into:
        \subsubsection{Single-Space Search}
             In this sub-category, both \ac{NN} candidates and compiler optimization possibilities share a unified search space, employing the same representation for both types of searched elements. In the unified search space, for each \ac{NN} operator, a list of possible compiler transformations is associated. The transformations can be architecture-related and changes the hyper-parameters such as number of groups, dilation rate in a convolution 2D, or schedule-related such as loop tiling or unrolling. The search process, as indicated in Figure~\ref{fig:taxonomy} (2.A), iteratively samples candidate couples from this unified search space and evaluates them for accuracy and efficiency.
             
             For instance, in~\citep{nasas}, \ac{NN} operators are represented in the same way as code transformations, i.e. as nested loops. These transformations are represented in the polyhedral model~\citep{polyhedral} and implemented in the TVM compiler~\citep{tvm}.
             The method replaces the convolution operators by trying out different sequences of transformations present in the search space, leveraging their inherent nested loop nature. The goal is to maintain comparable accuracy while reducing latency.
             AutoTVM~\citep{autotvm}, the \ac{ACO} module of TVM, is employed to perform auto-parameter-tuning on these transformations. To reduce the number of candidates, the search space is pruned by only keeping candidates that pass the Fisher Potential~\citep{fisher} test. Among the retained candidates, the best performing one is selected after the evaluation process, which is done by execution. Fisher Potential serves as a cost-effective zero-shot metric that rejects damaging network changes and transformations without training them. It is noteworthy, though, that despite its computational efficiency, this metric is proved to be sub-optimal~\citep{fisher-suboptimal}.

        \subsubsection{Dual-Space Search}
             In this sub-category 
             \ac{NN} architectures and compiler schedule candidates, while part of the same search process,  are defined in their distinct search spaces, $A$ and $S$ respectively.
             As illustrated in Figure~\ref{sec:search} (2.B), a (architecture operator, schedule) tuple candidate is sampled for accuracy and efficiency evaluation at each iteration. The search process is hindered by the large combined search spaces. 
             
             A notable work within this sub-category is NAAS~\citep{naas}. NAAS proposes an evolution-based framework that combines the architecture search process with compiler-level optimizations and hardware accelerator design. An accelerator is a hardware design specifically tailored to speed up certain computational tasks, such as convolution operations in \acp{NN}. The considered compiler transformations in this framework are Loop Ordering and Loop Tiling, which are used to optimize each convolution layer independently during the search. Based on these transformations, candidate optimizations proposed by the compiler are encoded as vectors assigning an order and a tile size to each convolution dimension (e.g., spatial width, kernel height, input channel). Accelerator design elements and compiler optimizations are integrated with the search space of the HW-NAS method called Once-For-All (OFA)~\citep{onceforall}, using the evolutionary algorithm.

            First, a pool of accelerator candidates is generated, and for each accelerator candidate, a network candidate that satisfies the pre-defined accuracy requirement is sampled from the OFA space. Then, for each (architecture, accelerator) pair, the algorithm searches for the optimal corresponding compiler optimization strategy. In order to take both latency and energy into consideration during the search, the Energy-Delay Product (EDP)~\citep{edp} is used as a minimization objective. The framework outputs a tuple of neural architecture, compiler schedule, and accelerator architecture. The accelerator architecture refers to the specific design and configuration of the hardware accelerator, which is optimized to maximize efficiency and performance.

\setlength\rotFPtop{0pt plus 1fil} 
\begin{sidewaystable}
\caption{Summary of existing NACOS methods and their characteristics}
\label{tab:methods}
\renewcommand{\arraystretch}{1.5}
\centering
\begin{tabular}{|>{\centering\arraybackslash}p{1cm}|>{\centering\arraybackslash}p{1cm}|>{\centering\arraybackslash}p{3cm}|>{\centering\arraybackslash}p{1.5cm}|>{\centering\arraybackslash}p{3cm}|>{\centering\arraybackslash}p{3cm}|>{\centering\arraybackslash}p{2cm}|>{\centering\arraybackslash}p{1.5cm}|}
\hline
\textbf{Method} & \textbf{Type} & \textbf{Target Hardware} & \textbf{Target NN} & \textbf{Objectives} & \textbf{Target Task} & \textbf{Open-Source} & \textbf{Compiler} \\ \hline
\citep{mcunet} & 1.A & MCU & CNN & Latency, memory \& storage usage, accuracy & Image Classification, Speech Recognition & Yes & TinyEngine \\ \hline
\citep{nasas} & 2.A & mCPU, mGPU, CPU, GPU & CNN & latency & Image Classification & Yes & TVM \\ \hline
\citep{sr} & 1.B & mGPU, mDSP & SRCNN & Latency, PSNR & Super Resolution & Yes & XGen \\ \hline
\citep{chanas} & 1.B & Mobile, CPU, GPU & CNN & Accuracy, latency & Image Classification & No & TVM \\ \hline
\citep{naas} & 2.B & NVDLA, Shidiannao, Eeyeriss, EdgeTPU & CNN & accuracy, EDP & Image Classification & No & MAESTRO \\ \hline
\end{tabular}
\end{sidewaystable}

\section{Search Strategies and Evaluation}~\label{sec:search}

Given the large size of search spaces in NACOS, in terms of \ac{NN} architectures and compiler schedules, it is important to use well-tailored exploration algorithms. Depending on the method, we can opt for a unified exploration algorithm to search for both \ac{NN} design units (operators, blocks or architectures) and compiler schedules simultaneously, or use distinct algorithms for each.

The most used strategy in this context is the Evolutionary Algorithm~\citep{evo}. It can be used for both One-stage and Two-Stage search techniques. In NAAS~\citep{naas}, it is used to sample (architecture, schedule) tuples for each explored hardware configuration. It can also be used for independently performing \ac{HW-NAS} like in MCUNet~\citep{mcunet}, or for applying \ac{ACO} like in CHaNAS~\citep{chanas}. 

Continuous optimization through gradient descent has also been used in~\citep{sr} by making the searched \ac{NN} hyper-parameters, such as the depth of the network and the per-layer width,  differentiable. 

A notable observation in two-stage search methods is that, it is common for the search strategy in the \ac{ACO} stage to be rule-based. For example, in~\citep{sr}, rules have been developed to perform different compiler optimizations, like fusing convolution operations with depth-to-space operations.

It is worth mentioning that some methods do not prioritize working on the search algorithm, but rather focus on designing and pruning the search space. That’s the case for~\citep{nasas}, where the search strategy is just enumerating random sequences of candidates from the search space and evaluating them.

Even with a well-crafted search space and an efficient exploration algorithm, it is unrealistic to measure the actual accuracy and hardware efficiency metrics during the search. Therefore, using performance and efficiency estimators becomes necessary to avoid training \ac{NN} candidates and executing them with sampled schedules at each search iteration. 

For accuracy, the dominating estimation strategy in NACOS methods~\citep{mcunet,chanas,naas} is One-shot evaluation, also referred to as weight sharing. It is often used when the architecture space $A$ is a super-network, and all possible candidate architectures are sub-networks of that super-network. In this method, the super-network is trained, then the sampled architectures, i.e sub-networks, are evaluated by inheriting the trained super-network weights, without having to train them separately. It is not the only one, though. In CHaNAS ~\citep{chanas}, a lookup table is used in the \ac{HW-NAS} stage and a Gaussian process-based cost model is used in the \ac{ACO} stage. In~\citep{sr}, a \ac{NN} speed model is used. Meanwhile, latency is measured through execution in~\citep{mcunet} and~\citep{nasas}. Memory footprint in MCUNet~\citep{mcunet} is also measured through execution.

\section{Challenges \& Limitations}~\label{sec:challenges}

In this section, we lay down the challenges and limitations present in current NACOS methods. We believe that addressing these issues is necessary for the progression of this research area, paving the way to unlock its full potential in generating optimal, efficient and hardware-friendly \ac{NN} architectures.

\subsection{Hardware heterogeneity}
As shown in Table~\ref{tab:methods}, existing NACOS methods target a variety of hardware platforms, spanning CPUs, GPUs, DSPs, mobile phones, \ac{NN} accelerators, and even \acp{MCU}. While this versatility allows for the generation of \ac{NN} models that can run on a large set of hardware platforms , from the most powerful hardware to the smallest IoT devices, it also exhibits several challenges. 

A primary concern is that these methods are limited by their use of compilers that are specific to the targeted hardware platforms. For example, in the case of MCUNet, TinyEngine is exclusively compatible with \acp{MCU}. This restricts the applicability of each method in terms of the used configuration specifications and hardware platforms.

With the evolving set of hardware and the growing need for neural network transferability across different devices, addressing this challenge of hardware heterogeneity becomes imperative for the advancement of this research area.
One promising approach, is promoted by the development of the MLIR~\citep{mlir} infrastructure. Indeed, MLIR~\citep{mlir} strives to support any high-level Domain Specific Language or \ac{DL} framework and compatible with different hardware devices. This will ensure that the co-searched neural network and compiler schedule pairs become usable and applicable on a wide range of off-the-shelf hardware.


\subsection {Generalization}

All of the existing NACOS methods focus on generating and optimizing \ac{CNN} architectures. These methods address architectures from both groups of \acp{CNN}~\citep{hwnas-survey} : standard \acp{CNN} that use standard convolutions, and extended \acp{CNN} that incorporate special convolution variants such as depthwise and grouped convolutions.  Some methods even extend their scope to specific types of \acp{CNN}, such as \ac{SR} \acp{CNN}, which are designed for super-resolution tasks and feature the same convolution blocks arranged in a different pyramidal structure~\citep{sr}. 

The inclination towards targeting \ac{CNN} architectures can be attributed to the loop-nest structure and affine nature of convolution operations~\citep{nasas}, which makes searching for compiler schedules and applying code optimizations on them easier and more natural. This is mainly due to the availability of well-defined loop representations in compiler Intermediate Representations (IRs) and the plethora of modules that optimize loop nests for different compilers (e.g.~\citep{tiramisu}).

However, several \ac{NAS} works have emerged to explore the search space for capsule networks~\citep{nascaps}, transformers~\citep{nas-nlp}, and GANs~\citep{nas-gans}. Despite these strides in generalization, none of the currently available \ac{NAS} techniques designed for tasks other than Computer Vision incorporate compiler optimizations within their process. This motivates interesting future works in which compiler optimizations can improve hardware efficiency across various domains. This exploration could potentially enable specialized language models to run efficiently on hardware-constrained devices

\subsection{Large search space exploration}
One of the most significant challenges in NACOS methods is the exploration of large search spaces. With standard \ac{NAS} strategies, the search process requires substantial computational resources. This limiting factor forces methodologies to stick to tiny machine learning tasks such as image classification (IC) with small datasets like CIFAR-10. 

The scope of NACOS can significantly be increased with pre-search strategies to reduce the search space size. In MCUNet~\citep{mcunet}, authors employ a search-space pruning heuristic to reduce its size, based on the assumption that a search space accommodating higher FLOPs under the specified memory constraint can produce better models. 

Besides, the unified search space requires careful definition. The application of different transformation and order on the operator needs heuristically based strategies. Over-aggressive loop unrolling or inappropriate tiling can lead to increased code size, cache misses, or even degraded performance. Strategies that balance these aspects use insights from \ac{ACO} methods~\citep{autotvm}. 

\subsection{Lack of benchmarks}
In the context of \ac{NAS}, datasets containing architectures along with their corresponding accuracy and hardware metrics, are known as \ac{NAS} Benchmarks~\citep{nas-bench-101,nas-bench-201,hw-nas-bench}. They are used to compare the performance of various \ac{NAS} algorithms while serving as search spaces. 

A significant challenge in NACOS is the absence of such benchmarks to reproduce and compare existing methods. This is mainly due to their nature, as they focus more on designing the search space rather than working on the search strategy. There are multiple ways in which the joint space can be designed and explored by the search algorithm, forming the basis of the entire method. This diversity makes it difficult to standardize the search spaces. Additionally, as shown in Table~\ref{tab:methods}, these techniques target different hardware platforms and employ various compilers to perform compiler scheduling. Creating a benchmark would thus require measuring the performance of models across numerous hardware platforms and using a variety of compilers. The dataset would also need to include (architecture, schedule) pairs, which further complicate this task because of the infinite possibilities of applicable compiler schedules.

\section{Conclusion}~\label{sec:conclusion}

This survey explores the combination of two efficient deep learning techniques: Hardware-Aware Neural Architecture Search (HW-NAS) and Automatic Code Optimization (ACO). The combination aims to alleviate the sub-optimality issues observed when these techniques are performed independently. We conduct a literature review and identify key challenges and limitations posed in existing works. Additionally, we propose potential solutions and improvements to further advance this research area. Unlocking the full potential of this area holds promise for enhancing not only neural network design but also the underlying infrastructure with minimal human intervention.

\bibliographystyle{unsrtnat}
\bibliography{main}

\end{document}